%% file: main_arxiv.tex
\documentclass{article} 
\usepackage{iclr2025_conference,times}

\input{math_commands.tex}

\definecolor{linkc}{rgb}{0, 0.44, 0.74}
\definecolor{eqc}{rgb}{1, 0, 0}
\definecolor{newcitecolor}{rgb}{0,0.6,0}
\usepackage[pagebackref=false,breaklinks=true,colorlinks=True,urlcolor=eqc,citecolor=linkc,linkcolor=eqc,bookmarks=false]{hyperref}
\usepackage{url}
\usepackage{graphicx}
\usepackage{booktabs}
\usepackage{wrapfig}
\usepackage{amsmath}
\usepackage{xspace}
\usepackage{textgreek}
\usepackage{bm}
\usepackage{multirow}
\usepackage[accsupp]{axessibility}  
\usepackage{colortbl}
\usepackage{orcidlink}
\usepackage{footmisc}
\usepackage{algorithm}
\usepackage{algpseudocode}
\usepackage{amsmath}
\usepackage{amsfonts}
\usepackage{listings}
\usepackage{graphicx}
\usepackage{caption}

\definecolor{mygreen}{RGB}{34,139,34}
\definecolor{mylightblue}{RGB}{0,162,230}
\definecolor{deepyellow}{RGB}{255, 215, 0} 
\definecolor{nvidiagreen}{RGB}{118, 185, 0}

\newcommand{\model}{ElasticDiT\xspace}

\makeatletter
\def\blfootnote#1{\xdef\@thefnmark{}\@footnotetext{\scriptsize #1}}
\makeatother

\title{\model: Efficient Diffusion Transformers via Elastic Architecture and Sparse Attention for High-Resolution Image Generation on Mobile Devices}

\author{%
  Kunpeng Du,\space\space\space
  Haizhen Xie,\space\space\space
  Sen Lu,\space\space\space
  Lei Yu,\space\space\space
  Binglei Bao,\space\space\space
  Huaao Tang,\space\space\space
  Chuntao Liu,\space\space\space
  \\
  \textbf{
  \vspace{0.15cm}
  Hao Wu,\space\space\space
  Yang Zhao,\space\space\space
  Zhicai Huang,\space\space\space
  Heyuan Gao\space\space\space
  Zhijun Tu,\space\space\space
  Jie Hu$^\ddagger$,\space\space\space
  Xinghao Chen$^\ddagger$\space\space\space
  }
  \\
  $\ddagger$ Corresponding Author,\space\space\space
  \normalfont{Huawei Technologies.}\quad
}

\iclrfinalcopy 
\begin{document}
\maketitle

\begin{figure}[htbp]
	\centering
	\includegraphics[width=1.0\linewidth]{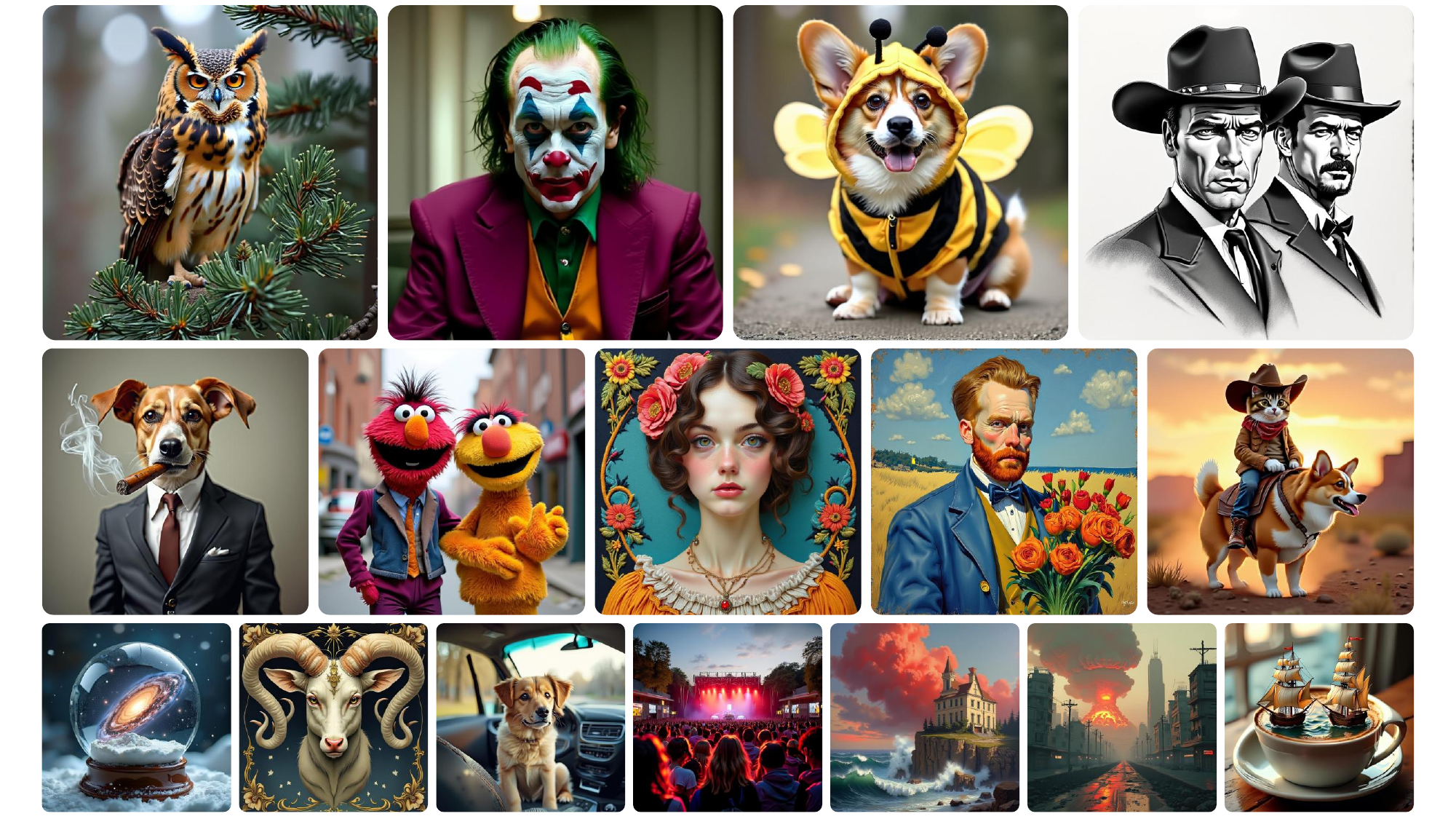}
	\caption{Generation results from our \model-flex-lite models.}
	\label{fig:fig1}
\end{figure}

\input{sections/0_abstract}

\begin{figure}[htbp]
	\centering
	\includegraphics[width=0.8\linewidth]{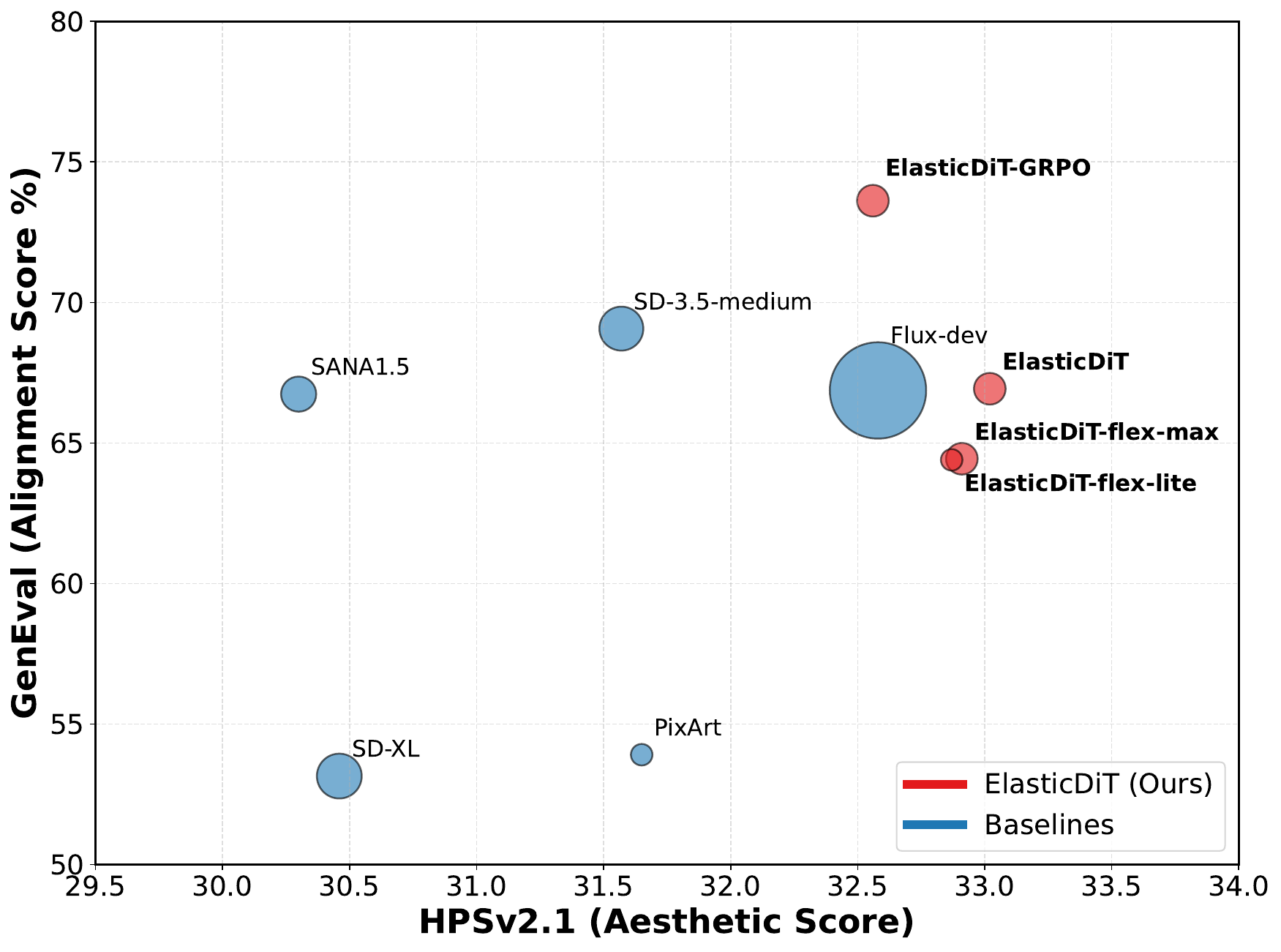}
	\caption{Comparison with other models on GenEval and HPSV2.1. Our models achieve the best trade-off of prompt adherence, image quality and model size.}
	\label{fig:fig1}
\end{figure}

\input{sections/1_introduction}

\input{sections/3_methods}

\input{sections/4_exps}

\input{sections/5_conclusion}


\clearpage

\bibliography{main}
\bibliographystyle{main}

\clearpage

\end{document}

%% file: math_commands.tex
\usepackage{amsmath,amsfonts,bm}

\def\eqref#1{equation~\ref{#1}}

\def\1{\bm{1}}

\def\vc{{\bm{c}}}

\def\vx{{\bm{x}}}

\DeclareMathAlphabet{\mathsfit}{\encodingdefault}{\sfdefault}{m}{sl}
\SetMathAlphabet{\mathsfit}{bold}{\encodingdefault}{\sfdefault}{bx}{n}



%% file: sections/0_abstract.tex
\begin{abstract}
    The Diffusion Transformer (DiT) architecture has emerged as the state-of-the-art paradigm for high-fidelity image generation, underpinning large-scale foundation models such as Stable Diffusion-3 and FLUX.1-dev for photorealistic synthesis. Deploying these models on resource-constrained mobile devices, however, entails prohibitive computational and memory overhead. While efficiency-driven approaches including Linear-DiT, DC-AE, and static pruning alleviate these bottlenecks, they frequently incur quality degradation, resulting in blurred details or structural artifacts. Unlike cloud environments where abundant resources permit hosting multiple specialized models, mobile memory constraints necessitate a versatile single-model paradigm that dynamically balances fidelity and latency. We therefore introduce ElasticDiT, which achieves this dynamic trade-off by adjusting both spatial compression ratios and DiT block depths. 
    By integrating Shift Sparse Block Attention (SSBA) and a Tiny DWT-Distilled VAE (T-DVAE), \model further reduces inference latency and memory footprint while maintaining comparable image quality. 
    Extensive experiments confirm that ElasticDiT effectively covers a wide range of fidelity-latency trade-offs within a single set of parameters. By jointly adjusting spatial compression ratios and DiT block depths, a single ElasticDiT model can be reconfigured on-the-fly to outperform various task-specific baselines across diverse hardware constraints. Specifically, our felx-lite variant achieves an HPS of 32.87, surpassing the Flux model, while maintaining competitive quality at 84.16\% average sparsity through SSBA. Furthermore, the plug-and-play T-DVAE provides SD3-level reconstruction with only 1/8$\times$ the computational cost of standard VAEs, and Flow-GRPO significantly boosts semantic alignment (GenEval: 66.93 $\rightarrow$ 73.62). These results demonstrate that ElasticDiT offers a versatile, hardware-adaptive solution that eliminates the need for multiple specialized models, providing a promising evolutionary path for future high-resolution image generation on mobile devices.
\end{abstract}

%% file: sections/1_introduction.tex
\section{Introduction}
\label{sec:intro}

The Diffusion Transformer (DiT) \cite{peebles2023scalable} has become the dominant architecture for high-fidelity image synthesis, as demonstrated by the success of large-scale models such as SD3 \cite{esser2024sd3} and Flux \cite{FLUX}. Despite their impressive generative performance, the substantial computational requirements and memory footprint of these models pose significant challenges for deployment on resource-constrained mobile devices. In particular, the inherent conflict between maintaining high-resolution output (e.g., 1024$\times$1024) and achieving low inference latency remains a critical bottleneck for on-device applications.

To mitigate these computational bottlenecks, recent research has pursued various efficiency-oriented strategies. For instance, Linear-DiT \cite{zhu2024dig} linearizes attention complexity through kernel-based approximations, while DC-AE \cite{chen2024deep} and SC-VAE \cite{scvae2025} employ high-ratio latent compression, and static pruning techniques \cite{snapegen2024} remove structural redundancies to reduce parameter counts. Despite their respective gains in inference speed, these approaches are fundamentally constrained by a deterministic architectural design, which binds the model to a fixed operating point within the fidelity-efficiency trade-off space. Such architectural rigidity is inherently incompatible with dynamic operational states and diverse task requirements. In practice, execution priorities frequently shift between efficiency-oriented inference and fidelity-centric synthesis; however, a static model lacks the flexibility to adapt its execution graph to these varying resource constraints and performance targets. Furthermore, the deployment of multiple task-specific variants remains prohibited by the stringent memory limits of mobile platforms. 

In this work, we introduce \model, a hardware-adaptive framework that addresses the fundamental trade-off between generative fidelity and inference latency via architectural elasticity. Unlike conventional static models, \model is built upon a unified Spatio-Depth Elastic Architecture, where a single set of parameters supports diverse execution states with varying structural granularities. By treating spatial compression ratios and transformer block depths as reconfigurable dimensions, the model can adapt its execution graph at runtime to align with instantaneous hardware constraints. This approach eliminates the storage overhead of maintaining multiple specialized checkpoints while enabling seamless transitions between efficiency-oriented and quality-centric synthesis.

To further enhance the efficiency of this elastic framework, we integrate a suite of synergetic optimizations tailored for mobile deployment. Specifically, we propose \textbf{Shift Sparse Block Attention (SSBA)} to address the quadratic complexity of standard self-attention. By partitioning 2-D latent spaces into sparse blocks and employing a window-shift strategy with layer-importance-aware execution, SSBA achieves near-linear complexity and sustains 84.16\% average sparsity while maintaining structural integrity. Additionally, we develop \textbf{Tiny Dwt-Distilled VAE (T-DVAE)}, which leverages Multi-level Haar Wavelet Transforms within a two-stage distillation framework. T-DVAE delivers reconstruction quality comparable to SD3-level VAEs at only 1/8$\times$ the computational cost, effectively mitigating the bottleneck of high-resolution latent decoding.
Experimental results demonstrate that \model offers a versatile and scalable solution for mobile generative tasks. Our flex-lite variant achieves competitive HPS scores compared to the 12B Flux model with much fewer parameters. By integrating Flow-GRPO for improved semantic alignment, ElasticDiT provides a viable architectural paradigm for future high-resolution image generation on mobile devices, bridging the gap between generation capacity and on-device efficiency.

%% file: sections/3_methods.tex
\section{Methods}

\begin{figure}[htbp]
	\centering
	\includegraphics[width=1.0\linewidth]{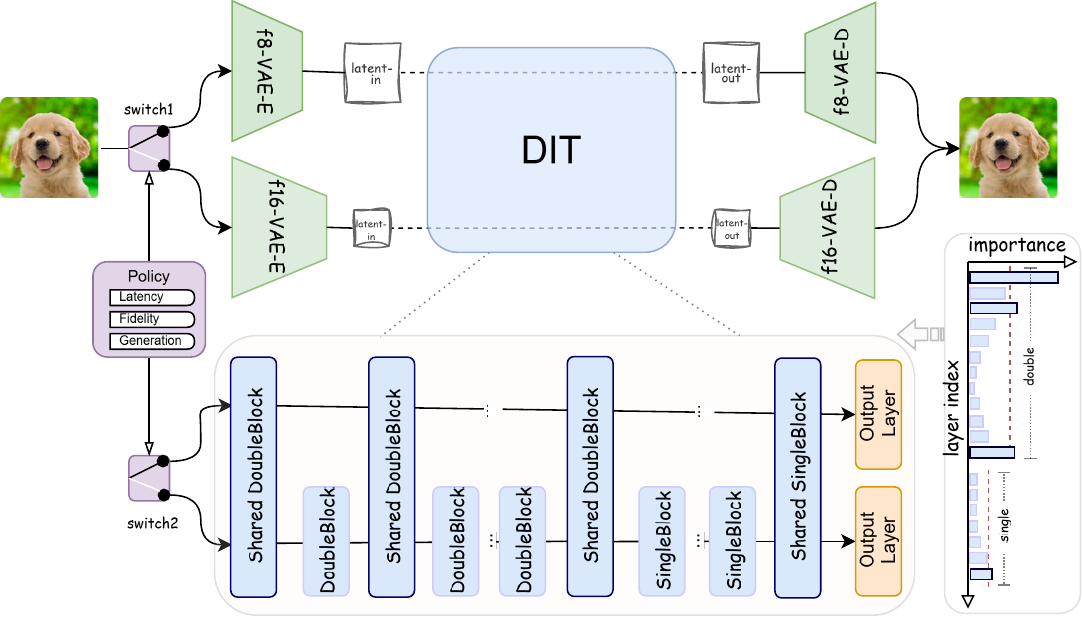}
	\caption{spatio-depth elastic architecture overview.}
	\label{fig:architecture23}
\end{figure}

\subsection{SPATIO-DEPTH ELASTIC ARCHITECTURE}

This novel framework is engineered as a unified, elastic weight repository that offers multi-dimensional scalability. Unlike static miniaturization, our approach enables on-the-fly re-configuration between heterogeneous VAE spatial compression ratios and varying DiT depths. This spatio-depth elastic architecture (Figure \ref{fig:architecture23}) treats inference latency as a dynamic constraint rather than a fixed parameter, reconciling the fidelity-efficiency trade-off through three synergistic mechanisms:
1) Dynamic VAE Routing, providing spatial elasticity by modulating latent resolution for instant throughput adjustment; 
2) Sparse-Depth Pruning, which provides depth elasticity by strategically bypassing redundant residual blocks from the full-backbone to accommodate varying computational budgets; and 3) Unified Weight Co-Optimization, which enforces parameter consolidation across the DiT-flex-max and DiT-flex-lite.

\subsubsection{Dynamic VAE Routing}
The original $\text{DiT}$ model suffers from quadratic computational complexity $O(N^2)$ in self-attention operations, leading to significant bottlenecks for high-resolution image generation. Our solution incorporates a Dual VAE Design consisting of two encoders: $\text{f8-VAE-E}$ and $\text{f16-VAE-E}$ (see Figure \ref{fig:architecture23}). Both produce 16-channel latent representations but with different spatial compression rates ($8\times$ vs $16\times$). For a given input resolution, the f16-VAE-E compresses the conditional image into smaller-sized latent features ($\text{e.g., } H/16 \times W/16$ instead of $H/8 \times W/8$) compared to the $\text{f8-VAE-E}$. This reduction in feature size $\mathbf{(H \cdot W \rightarrow H/2 \cdot W/2)}$ substantially reduces subsequent $\text{DiT}$ computation, leading to significant inference acceleration ($\approx 4\times$). However, this comes at the cost of reduced fidelity in the final generated image relative to the conditional input. We further introduce a lightweight Policy Network that dynamically selects the appropriate $\text{VAE}$ path ($\text{switch1}$) based on the desired fidelity-efficiency trade-offs.

\subsubsection{Sparse-Depth Pruning}
The Sparse-Depth Pruning mechanism is designed to support the switching between the DiT-flex-max and DiT-flex-lite using a shared set of weights, thereby managing the fundamental latency-fidelity capability trade-off. This system is constructed by first conducting layer importance analysis in our DiT-flex-max (illustrated in Figure \ref{fig:architecture23}). This analysis guides the construction of the efficient sub-network: DiT-flex-lite is formed by directly skipping $\text{DiT}$ blocks deemed to have relatively low importance. This mechanism proportionally reduces the number of blocks required for inference, directly achieving $\text{DiT}$ inference latency reduction based on the skipping ratio. Crucially, the reduction in model depth comes at the cost of reduced generative capability compared to the full $\text{DiT}$ backbone. The selection of the operational mode is managed by $\text{switch2}$ (see Figure \ref{fig:architecture23}) to facilitate the runtime adjustment of the latency-generative capability trade-off. To enhance fine-grained generation quality, we incorporate a zero-initialized $3 \times 3$ convolutional layer after the output projection, refining high-frequency details without extra inference cost.

\subsubsection{Unified Weight Co-Optimization}
\begin{figure}[htbp]
    \centering
    \includegraphics[width=0.7\linewidth]{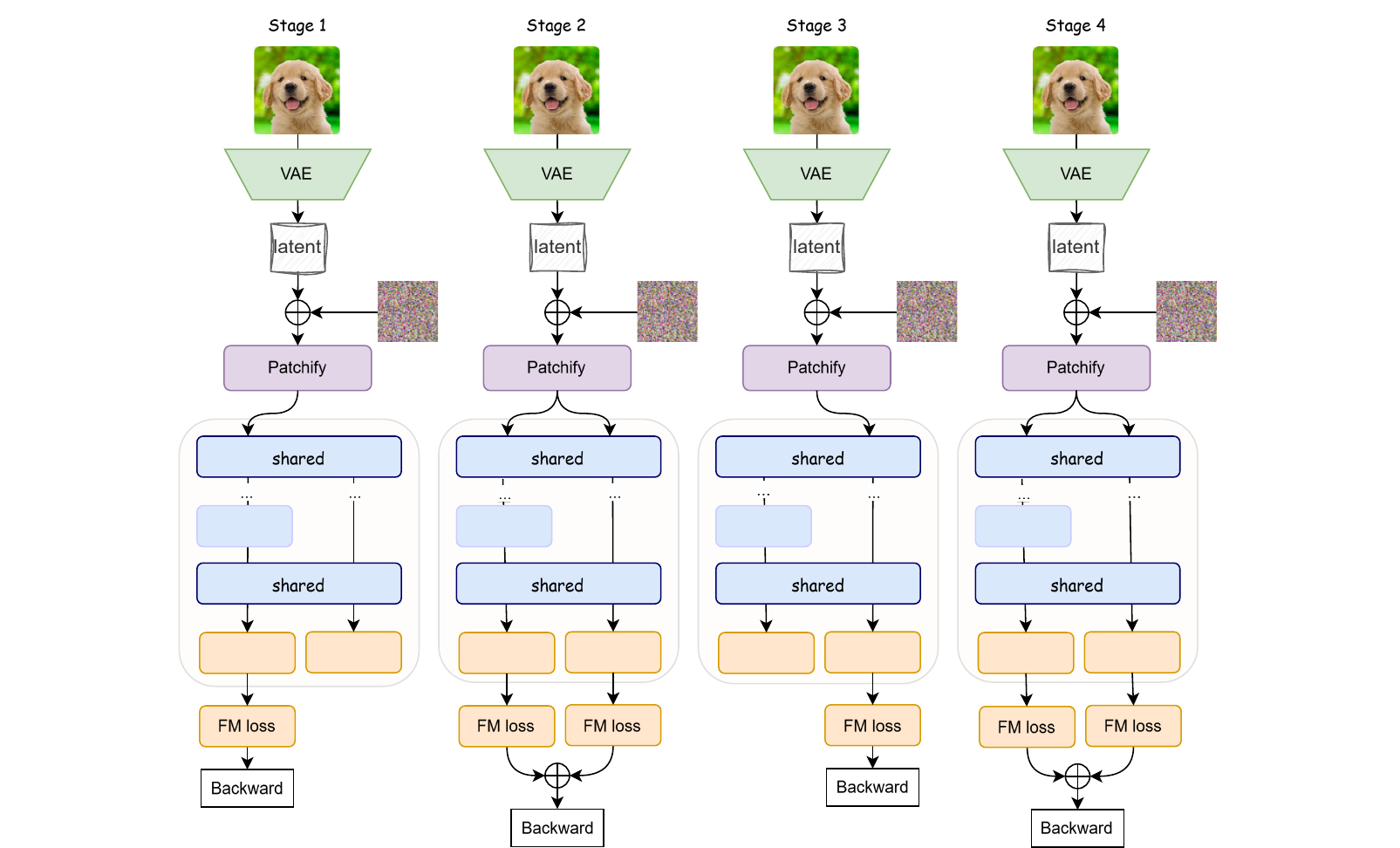}
    \caption{Unified Weight Co-Optimization.}
    \label{fig:elastic_train}
\end{figure}

To achieve full weight-sharing between the DiT-flex-max and DiT-flex-lite while ensuring high-quality generation capability, we design a rigorous Unified Weight Co-Optimization (UWCO) strategy. This co-training process (illustrated in the Figure \ref{fig:elastic_train}) comprises four distinct and sequential phases:

\begin{enumerate}
    \item \textbf{DiT-flex-max init:} The full-backbone DiT model is initially trained to convergence, establishing the upper bound for performance.
    \item \textbf{Hybrid Warm-up:} Both the DiT-flex-max and DiT-flex-lite models are jointly optimized on an identical data mixture. This phase is crucial for equipping the smaller, parameter-shared network with basic generative capacity.
    \item \textbf{DiT-flex-lite Specialisation:} Only the DiT-flex-lite path is activated and fine-tuned for $20\text{k}$ steps. This specialization phase targets the elimination of the performance gap between the resource-optimized variant and the full backbone.
    \item \textbf{Unified Recovery:} The complete model is re-enabled, and both the full and skip-layer paths are co-tuned, ensuring the final set of weights is robust across all configurations.
\end{enumerate}

Following these four stages, the system successfully stores only a single set of weights on disk. This enables deployment flexibility, allowing the model to be instantly configured either as a DiT-flex-max high-quality generator or as a DiT-flex-lite fast generator by simply toggling the skip-layer flag.

\subsection{SHIFT SPARSE BLOCK ATTENTION}

\begin{figure}[htbp]
	\centering
	\includegraphics[width=1.0\linewidth]{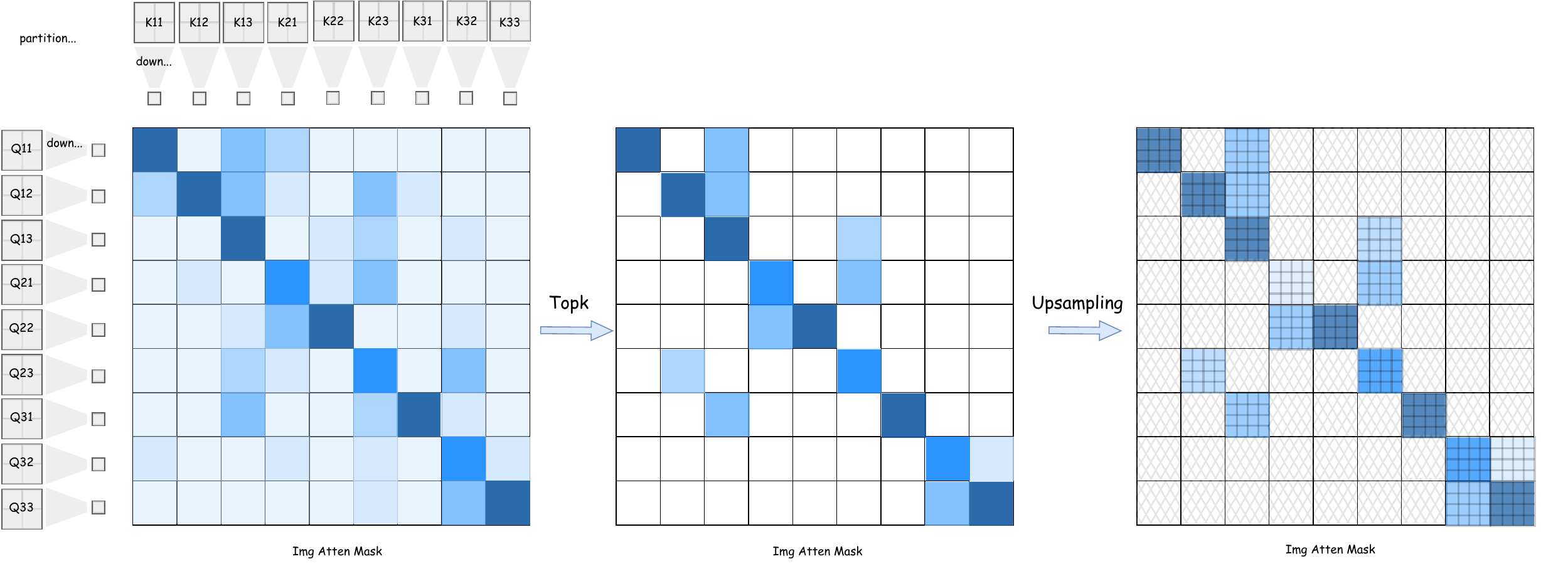}
	\caption{the sparse-block attention module that enables linear-complexity long-range interaction.}
	\label{fig:architecture}
\end{figure}

Modern text-to-image diffusion models rely on cascaded downsampling pipelines to alleviate computational burden. 
A typical \emph{f8c16} Variational Auto-Encoder (VAE) first compresses an RGB image $\mathbf{x}\in\mathbb{R}^{3\times 512\times 512}$ into a latent representation $\mathbf{z}\in\mathbb{R}^{16\times 64\times 64}$, yielding an $8\times$ spatial reduction. 
Subsequent $2\times 2$ patchification folds the latent tensor into a sequence of $1024$ image tokens, each of dimension 64. 
When text tokens are concatenated, the overall sequence length $T=T_{\text{img}}+T_{\text{txt}}$ becomes prohibitive for standard self-attention, whose complexity is $\mathcal{O}(T^{2})$. 
This quadratic scaling is infeasible for ultra-high-resolution scenarios, where the image token count alone can exceed several thousands.

\subsubsection{Sparse Block Attention}
To retain efficiency without sacrificing expressiveness, we propose a Sparse Block Attention (SBA) mechanism that adapts the Mixture of Block Attention (MoBA)~\cite{lu2025mobamixtureblockattention} paradigm to 2-D latent spaces. Sparse Block Attention is illustrated in the right part of Fig.~\ref{fig:architecture}.

\noindent \textbf{Block Compression.}
Given image tokens $\mathbf{Q_{\text{img}}}, \mathbf{K_{\text{img}}} \in \mathbb{R}^{T_{\text{img}} \times C}$ extracted from the latent representation, we partition them into non-overlapping $8\times 8$ blocks along the spatial dimensions, yielding $T_b = T_{\text{img}}/64$ blocks. Within each block, we compute the mean of the 64 token-wise queries and keys, producing compressed representations:
\begin{equation}
	\mathbf{q}_b = \text{MeanPool}_{8\times8}(\mathbf{Q_{\text{img}}}), \quad
	\mathbf{k}_b = \text{MeanPool}_{8\times8}(\mathbf{K_{\text{img}}}) \in \mathbb{R}^{T_b \times C}.
\end{equation}
Notably, the values $\mathbf{V} \in \mathbb{R}^{T_{\text{img}} \times C}$ remain at full resolution to preserve fine-grained local details during attention computation.

\noindent \textbf{Affinity Matrix.}
Leveraging the bidirectional attention pattern essential for diffusion denoising, we compute the affinity matrix between all compressed query and key blocks $\mathbf{S} = \mathbf{q}_b \mathbf{k}_b^{\top} \in \mathbb{R}^{T_b \times T_b}$.
This block-level affinity serves as a computationally efficient proxy for estimating token-level attention patterns, enabling global receptive field analysis with minimal overhead.

\noindent \textbf{Top-$k$ Gating.}
To drop the least-relevant blocks, we select the Top-$k$ smallest entries in $\mathbf{S}$ (where $k = \lfloor T_b \cdot \rho \rfloor$ and $\rho$ is the sparse ratio, typically 0.75) and mask the corresponding positions with $\mathbf{0}$, producing a gate matrix $\mathbf{G}\in\{0,1\}^{T_{b}\times T_{b}}$.
The mask is expanded to full token resolution by repeating each block entry to its corresponding $8\times 8$ token region, yielding $\mathbf{M}\in\{0,1\}^{T_{\text{img}}\times T_{\text{img}}}$. To ensure text tokens remain fully accessible, we pad $\mathbf{M}$ with ones along both axes for the text segment: every image query can attend to all text keys, while text queries attend to all text keys and sparsely to image keys.

\noindent \textbf{Efficient Sparse Execution.}
To leverage hardware-accelerated attention, we first cast $\mathbf{M}$ to a Boolean mask where \texttt{False} marks the pruned positions.
We emphasize that mere sparsification of the attention matrix does not reduce runtime latency; measurable speed-up is achieved only when vendor-supplied kernels explicitly elide the zeroed blocks during load/store operations.
Under this condition, the overall complexity drops from $\mathcal{O}(T^2)$ to $\mathcal{O}(T_b^{2}+T_{\text{img}}\cdot k\cdot 64)$, where $T_b^{2}$ accounts for block-level affinity computation and $T_{\text{img}}\cdot k\cdot 64$ corresponds to the sparse attention exerted on the selected blocks.
Algorithm~\ref{alg:sba} summarizes the complete procedure.

\begin{algorithm}[t]
	\caption{Sparse Block Attention}
	\label{alg:sba}
	\begin{algorithmic}[1]
		\Require Image tokens $\mathbf{Q}_{\text{img}},\mathbf{K}_{\text{img}},\mathbf{V}_{\text{img}} \in \mathbb{R}^{T_{\text{img}} \times C}$, text tokens $\mathbf{Q}_{\text{txt}},\mathbf{K}_{\text{txt}},\mathbf{V}_{\text{txt}} \in \mathbb{R}^{T_{\text{txt}} \times C}$, sparse ratio $\rho$
		\State Apply Rotary Position Embedding to $\mathbf{Q}_{\text{img}},\mathbf{K}_{\text{img}}$
		\State Partition image tokens into $8\times 8$ blocks: $T_b \gets T_{\text{img}}/64$
		\State $\mathbf{q}_b \gets \text{MeanPool}_{8\times8}(\mathbf{Q}_{\text{img}}), \quad \mathbf{k}_b \gets \text{MeanPool}_{8\times8}(\mathbf{K}_{\text{img}})$
		\State $\mathbf{S} \gets \mathbf{q}_b \mathbf{k}_b^{\top} \in \mathbb{R}^{T_b \times T_b}$ \Comment{Block-level affinity}
		\State $k \gets \lfloor T_b \cdot \rho \rfloor$ \Comment{Dynamic sparse selection}
		\State $\mathbf{G} \gets \text{Top-$k$Smallest}(\mathbf{S}, k)$ \Comment{Select least relevant blocks}
		\State $\mathbf{M}_{\text{img}} \gets \text{ExpandMask}(\mathbf{G}, 8\times8)$ \Comment{Expand to token resolution}
		\State $\mathbf{M} \gets \text{Pad}(\mathbf{M}_{\text{img}}, T_{\text{txt}}, \text{axis}=(0,1))$ \Comment{Text tokens fully visible}
		\State $\mathbf{Q} \gets [\mathbf{Q}_{\text{txt}}; \mathbf{Q}_{\text{img}}], \quad \mathbf{K} \gets [\mathbf{K}_{\text{txt}}; \mathbf{K}_{\text{img}}], \quad \mathbf{V} \gets [\mathbf{V}_{\text{txt}}; \mathbf{V}_{\text{img}}]$
		\State $\mathbf{O} \gets \text{SparseAttention}(\mathbf{Q}, \mathbf{K}, \mathbf{V}, \mathbf{M})$ \Comment{Hardware-optimized kernel}
		\State \Return $\mathbf{O}[T_{\text{txt}}:]$ \Comment{Return image features}
	\end{algorithmic}
\end{algorithm}

\subsubsection{Window Shift Strategy}
\label{sec:swin_shift}

While SBA achieves significant complexity reduction, the fixed $8\times 8$ block partitioning may introduce grid artifacts near object boundaries and fine-grained textures.
Inspired by Swin Transformer~\cite{liu2021swin}, we enhance SBA with a Shift Window Strategy that alternates between two grid configurations, breaking artificial partition boundaries while maintaining sparse computation patterns.
Given a latent map of spatial dimensions $H\times W$ (typically $64\times 64$), at layer $\ell$ we alternate between two offset vectors:
\[
\Delta^{(\ell)} = 
\begin{cases}
	(4,4) & \text{\texttt{if $\ell$ \% 2 == 0}} \\
	(-4,-4) & \text{\texttt{else}}
\end{cases}
\]
The grid origin is shifted by $\Delta^{(\ell)}$ before block partitioning, effectively displacing each $8\times 8$ window diagonally by half its size in alternating layers.
We implement this efficiently using \texttt{torch.roll} with zero memory overhead on modern accelerators.

\subsection{Tiny DWT-Distilled Vae}
\begin{figure}[htbp]
	\centering
	\includegraphics[width=1.0\linewidth]{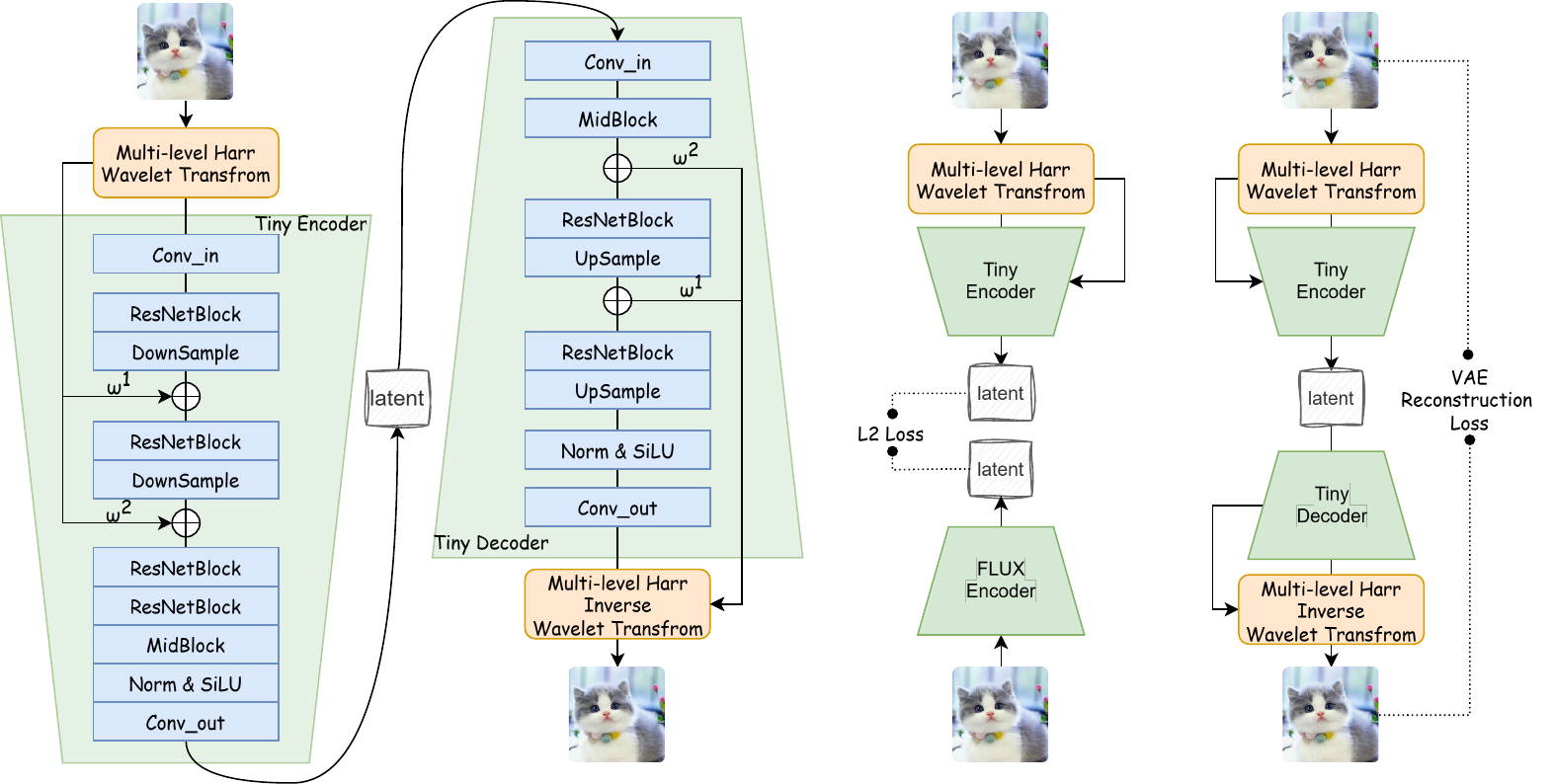}
	\caption{T-DVAE pipeline. \textbf{Left:} encoder compatible with \model, enhanced by multi-level Haar wavelets. \textbf{Right:} two-stage distillation — latent alignment and reconstruction training.}
	\label{fig:vae}
\end{figure}

Traditional VAE such as FLUX-VAE~\cite{esser2024scaling} has excellent image reconstruction capabilities, while its enormous computing requirements limit its application in edge computing scenarios. In order to reduce the training cost and inference latency of text-to-image generation, we design an efficient lightweight VAE, which reduces the computational load by using multi-level wavelet transform like WF-VAE \cite{chen2024wfvae} and ensures high reconstruction quality via two-stage distillation training paradigm. 

\subsubsection{DWT-Enhanced Architecture}
The overall architecture is displayed in Fig.~\ref{fig:vae} left.
A single 2-D Haar kernel bank
$\mathcal{F}=\{\psi_{\text{LL}},\psi_{\text{HL}},\psi_{\text{LH}},\psi_{\text{HH}}\}$
is recursively injected after each stride-2 conv:
\begin{equation}
	\mathcal{W}^{(l)}=\mathcal{W}^{(l-1)}_{\text{c}}*\mathcal{F},\quad
	\mathcal{W}^{(l)}\in\mathbb{R}^{4C\times\frac{H}{2^l}\times\frac{W}{2^l}},
\end{equation}
where $\text{c}=\{\text{LL},\text{HL},\text{LH}\}$ keeps the low-frequency bulk plus a slice of high-frequency edges.
The decoder symmetrically inverts the pyramid:
\begin{equation}
	\hat{\mathcal{W}}^{(l-1)}_{\text{c}}=\text{IWT}\bigl(\mathcal{W}^{(l)}\bigr)+\Delta^{(l-1)}_{\text{c}},\quad
	\hat{I}=\text{IWT}\bigl(\mathcal{W}^{(0)}\bigr).
\end{equation}

To maintain the compression ratio, we removed one upsampling layer and one downsampling layer from the original structure.
This modification halved the computational resolution of the VAE, effectively reducing the overall computational load.
Additionally, we further pruned the remaining VAE structure by removing the mid-block attention, which has a minimal impact on frequency domain reconstruction capabilities, and replacing several deep resblocks with grouped convolutions, further reducing the number of parameters and computational load.

\subsubsection{Two-Stage Latent Distillation}
During the training phase, to improve the model's reconstruction capabilities, we proposed the Tiny-VAE training paradigm based on DC-AE, dividing the entire VAE training process into two stages.
In the first stage, we only activate the encoder and use a simple L2-loss to align the latent encoded by FLUX encoder, which can be expressed by:
\begin{equation}
	L_{Encoder} = L_{l2}(E_{flux}(I), E(I)).
\end{equation}
In the second stage, we fix the encoder and activate the decoder to train the reconstruction capability of the VAE after latent domain alignment, using standard VAE reconstruction losses, including L2-loss, perceptual loss, and patch-GAN loss.
The training objective of the decoder is:
\begin{equation}
	L_{\text{Decoder}} = \lambda_1 \cdot L_{\text{l2}}(I, D(E(I))) + \lambda_2 \cdot L_{\text{perceptual}}(I, D(E(I))) + \lambda_3 \cdot L_{\text{patchGAN}}(I, D(E(I))).
\end{equation}
Practical results show that the VAE trained using this training paradigm achieves better reconstruction performance than the single-stage trained VAE, and the aligned latent domain facilitates the training of DiTs.
Additionally, to make the model easier to train, we employ a reparameterization technique, replacing the single convolution in the network with a multi-branch structure during training, increasing its trainable parameters, accelerating convergence, and improving reconstruction metrics.

\section{Efficient Training/Inference} 

\subsection{Data Curation and Training Pipeline} 
To support the training of a text-to-image diffusion model with high visual fidelity and semantic alignment, we first assembled $\sim$1B raw image--text pairs harvested from both open-source repositories and industrial partners.
We designed a multi-stage data-refinement protocol to maximize the signal-to-noise ratio of the training corpus:
\emph{Resolution \& aspect-ratio filtering:} discard images shorter than 512 px on either side or whose aspect ratio deviates by $>2\times$.
\emph{Aesthetic \& technical quality pruning:} retain only samples whose AES (Aesthetic Score) and LIQE (Learning-based Image Quality Evaluator) exceed dataset-specific thresholds $\tau_{\text{AES}}$ and $\tau_{\text{LIQE}}$ calibrated on a human-rated validation set; subsequently, IQA models (MUSIQ, HyperIQA) eliminate technically defective images (blur, over-exposure, JPEG artifacts).
\emph{Semantic re-alignment and anomaly suppression:} each remaining image is re-captioned by InternVL2-7B, a SOTA vision-language foundation model, producing longer, noun-rich descriptions;
Concurrently, VLM-based anomaly detectors flag anatomical implausibilities (distorted limbs, extra fingers, asymmetric faces) and exclude them from the training pool.

\subsection{Training and Inference} 
\noindent \textbf{Pre-training.}
The base model is optimized on $\sim$1B real image--text pairs augmented with 5M high-resolution samples curated from the HuggingFace Hub and 1M images synthesized by a proprietary FLUX-based diffusion pipeline, giving $\sim$1.06B effective training instances.
Following best practices for scalable diffusion training~\cite{rombach2022high,podell2023sdxl}, we adopt stratified sampling with an 8:2 real-to-synthetic ratio, which empirically accelerates convergence and improves generalization to out-of-distribution prompts without inducing synthetic-artifact overfitting.

\noindent \textbf{Supervised Fine-Tuning.}
We construct a 6M synthetic prompt--image gallery and subject it to an automated quality-assurance stack:
After filtering, $\sim$5.2M images remain. We pair them with 100K English prompts and 5K Chinese prompts sampled from our internal multilingual prompt bank to endow the model with cross-lingual generation capacity.
Collectively, these data-centric designs establish a robust foundation for subsequent high-resolution, prompt-adherent text-to-image synthesis.

\noindent \textbf{Reinforcement Learning. }
We align the model's output distribution with human preferences following Flow-GRPO~\cite{liu2025flow}, focusing on prompt adherence while maintaining the generation quality. The denoising process is an MDP trajectories, with a group of \(G\) individual images \(\{\vx^i_0\}_{i=1}^G\) sampled by the flow model \(p_{\theta}\) using a prompt \(\vc\). The GRPO algorithm optimizes the policy model by maximizing the following objective: 
\begin{equation}
	\mathcal{J}_\text{Flow-GRPO}(\theta) = \mathbb{E}_{\vc\sim \mathcal{C}, \{\vx^i\}_{i=1}^G\sim \pi_{\theta_\text{old}}(\cdot\mid \vc)} f(r,\hat{A},\theta, \varepsilon, \beta),
	\label{eq:grpoloss}
\end{equation}
where
\begin{equation}
	f(r,\hat{A},\theta, \varepsilon, \beta) =
	\frac{1}{G}\sum_{i=1}^{G} \frac{1}{T}\sum_{t=0}^{T-1} \Bigg( 
	\min \Big( r^i_t(\theta) \hat{A}^i,  
	\ \text{clip} \Big( r^i_t(\theta), 1 - \varepsilon, 1 + \varepsilon \Big) \hat{A}^i \Big)
	- \beta D_{\text{KL}}(\pi_{\theta} || \pi_{\text{ref}}) 
	\Bigg), \\
\end{equation}

The $r^i_t(\theta)$ stands for the probability ratio between the prediction of $p_{\theta}$ and that of $p_{\theta_{old}}$  for predicting $x_{t-1}$ given $x_t$ and $c$. The $\hat{A}$ is the group relative advantage calculated by generation rewards.

%% file: sections/4_exps.tex
\section{Experiments}

\subsection{Model Details}
Fig.~\ref{fig:architecture} details the architecture of \model.
To maximize parameter reuse and inherit the comprehensive prior knowledge of the larger model, \model-flex-lite is architecturally defined as a subset of \model-flex-max, as shown in the bottom of Fig.~\ref{fig:architecture23}, both models share an identical transformer block design and differ only in network depth.

\begin{table}[t]
    \centering
    \caption{Quantitative comparison of state-of-the-art T2I models at $1024\times1024$ resolution. Left: HPSv2.1 scores across four visual domains and their average. Right: GenEval fine-grained metrics and overall score. Our \model-flex-lite surpasses FLUX with fewer parameters, while \model-GRPO attains the best overall GenEval performance, demonstrating superior semantic alignment.}
    \label{tab:model_comparison}
    \resizebox{\linewidth}{!}{%
        \begin{tabular}{lccccccccccccc} 
            \toprule
            & & \multicolumn{5}{c}{\textbf{HPSv2.1}} & \multicolumn{7}{c}{\textbf{GenEval}} \\ 
            \cmidrule(lr){3-7} \cmidrule(lr){8-14} 
            \textbf{Model} & \textbf{Res.} & Photo & Animation & Concept & Paint & \textbf{Averaged} & two\_obj & position & single\_obj & counting & color\_attr & colors & \textbf{Overall Score} \\
            \midrule
            PixArt         & 1024 & 30.53 & 32.90 & 31.58 & 31.58 & 31.65 & 61.62\% & 12.50\% & 98.44\% & 45.00\% & 24.50\% & 81.38\% & 53.91\% \\
            SD-XL          & 1024 & 28.76 & 31.90 & 30.66 & 30.51 & 30.46 & 67.93\% & 11.25\% & 99.38\% & 38.12\% & 19.00\% & 83.24\% & 53.15\% \\
            SD-3.5-medium  & 1024 & 29.11 & 33.39 & 32.40 & 32.11 & 31.57 & 85.61\% & 24.00\% & 99.69\% & 65.94\% & 56.75\% & 82.45\% & 69.07\% \\
            FLUX.1-dev     & 1024 & 31.22 & 33.92 & 32.35 & 32.83 & 32.58 & 80.05\% & 24.50\% & 98.44\% & 70.31\% & 51.00\% & 76.86\% & 66.87\% \\
            SANA1.5-1.6B   & 1024 & 29.43 & 31.62 & 30.13 & 30.01 & 30.30 & 82.58\% & 21.50\% & 98.75\% & 63.12\% & 47.50\% & 86.97\% & 66.74\% \\
            \midrule
            \model    & 1024 & 31.78 & 34.28 & 32.91 & 33.10 & 33.02 & 80.56\% & 18.50\% & 100.00\% & 63.75\% & 55.25\% & 83.51\% & 66.93\% \\
            \model-GRPO    & 1024 & 31.22 & 33.95 & 32.46 & 32.63 & 32.56 & 92.17\% & 20.75\% & 99.69\% & 79.06\% & 66.00\% & 84.04\% & 73.62\% \\
            \midrule
            \model-flex-max & 1024 & 31.53 & 34.23 & 32.85 & 33.03 & 32.91 & 83.84\% & 13.25\% & 99.68\% & 67.19\% & 45.00\% & 77.66\% & 64.44\% \\
            \model-flex-lite & 1024 & 31.52 & 34.19 & 32.75 & 33.04 & 32.87 & 80.81\% & 13.00\% & 99.69\% & 65.94\% & 48.50\% & 78.46\% & 64.40\% \\
            \bottomrule
        \end{tabular}}
\end{table}

\begin{table}[t]
	\centering
	\caption{Quantitative comparison of baseline dense DiT, Vanilla-Sparse, and Layer-Aware-Sparse \model at 512$\times$512 resolution. Layer-Aware-Sparse, operating at 84.16\% average sparsity, achieves the best overall generation quality while substantially reducing inference cost.}
	\label{tab:tb2}
	\resizebox{\linewidth}{!}{%
		\begin{tabular}{lcccccccccccccc}
			\toprule
			& & & & \multicolumn{2}{c}{\textbf{HPSv2.1}} & \multicolumn{9}{c}{\textbf{GenEval}} \\ \cmidrule(lr){4-7} \cmidrule(lr){9-14}
			\textbf{Model} & \textbf{Sparsity} & \textbf{Res.} & Photo & Animation & Concept & Paint & \textbf{Averaged} & two\_obj & position & single\_obj & counting & color\_attr & colors & \textbf{Overall Score} \\
			\midrule
			baseline & 0\% & 512 & 30.11 & 32.73 & 31.18 & 31.68 & 31.42 & 75.25\% & 16.75\% & 99.06\% & 60.94\% & 46.00\% & 77.39\% & 62.57\% \\
			Vanilla-Sparse & 75.00\% & 512 & 30.10 & 32.58 & 31.08 & 31.59  & 31.32 & 75.25\% & 15.50\% & 98.75\% & 61.88\% & 45.75\% & 77.93\% & 62.51\% \\
			Layer-Aware-Sparse & 84.16\% & 512 & 30.12 & 32.64 & 31.12 & 31.59  & 31.36 & 75.00\% & 16.75\% & 99.06\% & 64.06\% & 45.00\% & 76.06\% & 62.66\% \\
			\bottomrule
	\end{tabular}}
\end{table}
\begin{table}[h]
	\centering
	\caption{Reconstruction comparison of VAE variants on \texttt{test\_2048}.}
	\label{tab:tb3}
	\begin{tabular}{lccccc}
		\toprule
		VAE model & PSNR$\uparrow$ & SSIM$\uparrow$ & LPIPS$\downarrow$ & FID$\downarrow$ \\
		\midrule
		F8-VAE & 39.78 & 0.9686 & 0.0211 & 0.43 \\
		F16-VAE & 33.39 & 0.8965 & 0.0745 & 1.98 \\
		\bottomrule
	\end{tabular}
\end{table}

\begin{table}[h]
	\centering
	\caption{Latency-quality trade-offs of the four plug-and-play DiT--VAE configurations. ``Latency'' denote per-stage running time; ``Fidelity'' indicate reconstruction or generative quality. Switch-1 selects the VAE precision (F8/F16), while Switch-2 chooses the model size (max/lite), enabling on-device deployment across real-time or high-quality modes.}
	\label{tab:tb4}
	\begin{tabular}{llcc}
		\toprule
		Strategy & Model & Latency & Fidelity \\
		\midrule
		\multirow{2}{*}{Switch 1}
		& F8-VAE  & High & High \\
		& F16-VAE & Low  & Low \\
		\addlinespace
		\multirow{2}{*}{Switch 2}
		& DiT-max & High & High \\
		& DiT-lite & Low  & Low \\
		\bottomrule
	\end{tabular}
\end{table}

\begin{table}[ht]
	\centering
	\caption{Reconstruction quality vs.\ computational cost of different VAEs at 1024$\times$1024.}
	\label{tab:tb5}
	\begin{tabular}{lcccccc}
		\toprule
		VAE model & Resolution & TFlops & PSNR$\uparrow$ & SSIM$\uparrow$ & LPIPS$\downarrow$ & FID$\downarrow$ \\
		\midrule
		Flux VAE & 1024 & 14.34 & 32.12 & 0.9470 & 0.0225 & 1.72 \\
		SD3 VAE & 1024 & 14.34 & 29.75 & 0.9168 & 0.0352 & 2.78 \\
		Ostris VAE & 1024 & 7.35 & 29.71 & 0.9130 & 0.0341 & 3.67 \\
		SD VAE & 1024 & 14.33 & 25.24 & 0.8146 & 0.0768 & 22.89 \\
		T-DVAE1 & 1024 & 3.37 & 31.01 & 0.9319 & 0.0278 & 2.25 \\
		T-DVAE2 & 1024 & 1.81 & 29.57 & 0.9137 & 0.0367 & 2.78 \\
		\bottomrule
	\end{tabular}
\end{table}


\begin{table}[t!]
	\centering
	\caption{Plug-and-play replacement of Flux-VAE with our distilled Tiny-VAE in a 512$\times$512 DiT trained on Flux latents. Top: HPSv2.1 and GenEval metrics without any re-finetuning. Tiny-VAE preserves HPS while boosting GenEval (62.49\% $\rightarrow$ \textbf{63.37\%}), validating effective latent-space alignment and robust generative transfer.}
	\label{tab:tb6}
	\resizebox{\linewidth}{!}{%
		\begin{tabular}{lccccccccccccc}
			\toprule
			& & & & \multicolumn{1}{c}{\textbf{HPSv2.1}} & \multicolumn{9}{c}{\textbf{GenEval}} \\ \cmidrule(lr){3-6} \cmidrule(lr){8-13}
			\textbf{Model} & \textbf{Res.} & Photo & Animation & Concept & Paint & \textbf{Averaged} & two\_obj & position & single\_obj & counting & color\_attr & colors & \textbf{Overall Score} \\
			\midrule
			Base (Flux VAE) & 512 & 29.14 & 31.96 & 30.30 & 30.89 & 30.57 & 79.80\% & 15.50\% & 98.75\% & 54.06\% & 46.25\% & 80.59\% & 62.49\% \\
			Base (T-DVAE) & 512 & 29.09 & 31.86 & 30.27 & 30.89 & 30.53 & 79.29\% & 16.00\% & 98.75\% & 56.56\% & 48.25\% & 81.38\% & 63.37\% \\
			\bottomrule
	\end{tabular}}
\end{table}

\subsection{Evaluation Details}
We evaluate the performance of our \model using the two most mainstream metrics, HPSv2 (Human Preference Score v2)~\citep{wu2023human} and GenEval~\citep{ghosh2024geneval}, and compare it with state-of-the-art (SOTA) methods.
HPSv2 is an automated metric that quantifies the human-perceived quality of text-to-image (T2I) generation. It includes 400 distinct prompts, with each prompt generating 9 images, resulting in a test set of 3,600 images. By contrast, GenEval focuses on text-image alignment that a core requirement for practical T2I systems. Its test set contains 1,065 curated prompts, designed to assess key alignment capabilities (e.g., fine-grained attributes, spatial relationships). With a multi-dimensional scoring framework, it robustly evaluates whether generated images faithfully reflect text semantics.

\subsection{Performance Comparison and Analysis}
Compared with SOTA methods (Tab.~\ref{tab:model_comparison}), \model-flex-lite achieved an HPS of 32.87, surpassing Flux-12B while using 20$\times$ fewer parameters, demonstrating superior generation quality. On GenEval, \model trailed SD-3.5-medium by a small margin but performed on par with Flux-12B, evidencing competitive semantic alignment. After applying GRPO to \model, the HPS slightly decreased (33.02 $\rightarrow$ 32.56) while GenEval substantially improved (66.93 $\rightarrow$ \textbf{73.62}), validating the effectiveness of the proposed reinforcement learning enhancement.

At 512$\times$512 resolution (Tab.~\ref{tab:tb2}), Vanilla-Sparse with 75\% uniform sparsity accelerated attention inference by over 2$\times$ without degrading HPS or GenEval scores. By further adopting the layer-importance-aware scheme, average sparsity increased to 84.16\%, yielding additional speed-up and establishing a new Pareto frontier between latency and quality, thus enabling T2I generation at 2K and beyond.

As shown in Tab.~\ref{tab:tb3}, F8-VAE surpassed F16-VAE on all reconstruction metrics, whereas F16-VAE reduced latent tokens to 25\% of the F8 setting, achieving 4$\times$ inference acceleration at the cost of minor quality loss. Coupling either VAE with \model-flex-max/lite produced four plug-and-play configurations whose latency–fidelity trade-offs are summarized in Tab.~\ref{tab:tb4}, facilitating flexible on-device deployment under real-time or quality constraints.

Against SOTA VAEs on 1024-resolution images (Tab.~\ref{tab:tb5}), the 500\,ms Tiny-VAE matched Flux-VAE in PSNR, SSIM and FID while outperforming SD3-VAE and Ostris-VAE; the 300\,ms variant achieved SD3-level quality with 2--3$\times$ lower latency, demonstrating that the two-stage distillation paradigm can substantially reduce computational overhead without sacrificing reconstruction fidelity.

Plugging Tiny-VAE into the 512$\times$512 DiT originally trained with Flux-VAE---no re-finetuning---yielded almost identical HPS and a marginal GenEval gain (62.49\% $\rightarrow$ \textbf{63.37\%}, Tab.~\ref{tab:tb6}), evidencing accurate latent-space alignment and robust generative transfer of the distilled VAE.

\begin{figure}[t!]
	\centering
	\includegraphics[width=0.92\linewidth]{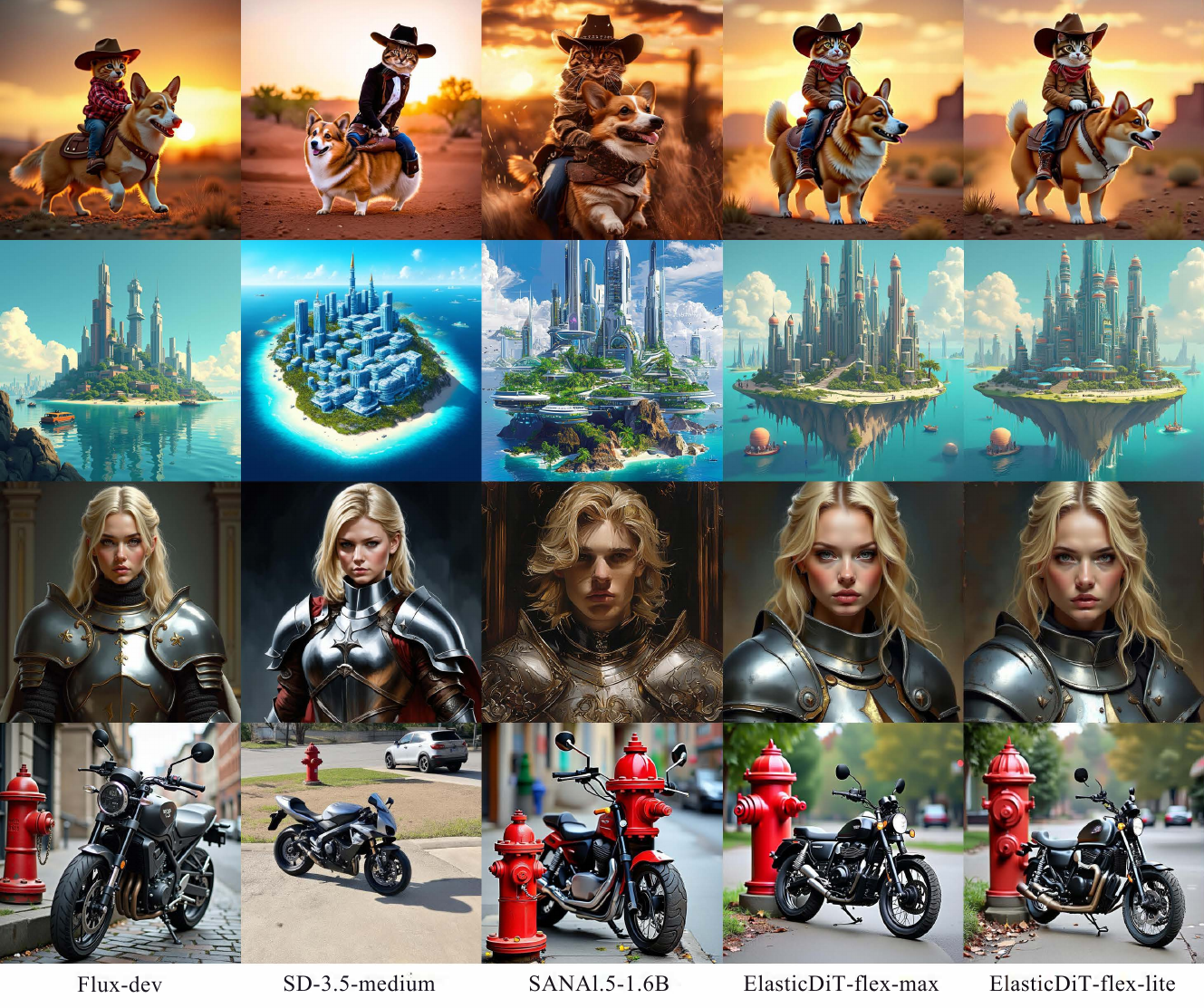}
	\caption{T2I qualitative comparison.}
	\label{fig:fig_ab}
\end{figure}

%% file: sections/5_conclusion.tex
\section{Conclusion}

This paper presents \model, a hardware-aware framework that significantly improves the quality-efficiency trade-off for mobile image generation. The proposed Spatio-Depth Elastic Architecture provides the flexibility to reconfigure model depth and resolution on-the-fly. This ensures optimal performance across varying hardware limitations with a unified set of parameters. The integration of SSBA and T-DVAE further augments the inference efficiency of \model, achieving linear complexity and efficient decoding while maintaining high reconstruction fidelity.
Experimental results confirm that our \model-flex-lite variant achieves performance on par with the much larger Flux model, representing a significant breakthrough in parameter efficiency. These findings establish ElasticDiT as a scalable and practical blueprint for high-resolution generation on edge devices. Future work will extend these principles to video generation and explore sparse Mixture-of-Experts (MoE) to further scale model capacity within limited computational budgets.